# Sentiment Analysis on Customer Responses

Antony Samuels, John Mcgonical

University of Southern California, Caltech

aantonysamuels@gmail.com  to.john.mcgonical@gmail.com

*Abstract*- Sentiment analysis is one of the fastest spreading research areas in computer science, making it challenging to keep track of all the activities in the area. We present a customer feedback reviews on product, where we utilize opinion mining, text mining and sentiments, which has affected the surrounded world by changing their opinion on a specific product. Data used in this study are online product reviews collected from Amazon.com. We performed a comparative sentiment analysis of retrieved reviews. This research paper provides you with sentimental analysis of various smart phone opinions on smart phones dividing them Positive, Negative and Neutral Behaviour.

## I. INTRODUCTION

Opinions are statements that reflect people's perception or sentiment. Sentiment analysis is a series of methods, techniques, and tools about detecting and extracting subjective information, such as opinion and attitudes, from language [1], helping in finding the mood of the customers about a purchasing of a particular product or topic. It involves building a system to collect and examine opinions about the product made in many online purchasing sites. Sentiment analysis is a sub field of web content mining. This paper is organized as follows: Section I Introduction, Section II Related Work, Section III Opinion mining and Sentimental analysis, Section IV Sentimental Classification, Section V Opinion Mining Techniques, and Section VI Tools Used in opinion mining, Section VII Applications, Section VIII Research challenges and in the last Section Research Scope.

## II. RELATED WORK

Conventionally, sentiment analysis has been about opinion contradiction, i.e., whether someone has positive, neutral, or negative opinion towards something [2]. Data used in this paper is a combination of product reviews collected from Amazon.com [3], between July and September, 2018. There have been somewhat overcome in the following two manners: Firstly, each product review holds inspections before it can be posted. Secondly, respective review must have a rating on it that can be used as the ground truth. The rating is based on a star scaled system, where the highest rating has 5stars and the lowest rating has only 1 star Figure1.

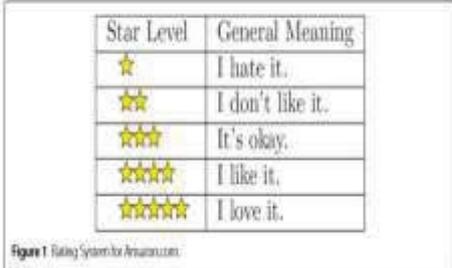

Figure 1 Rating System for Amazon.com

Here, the report tackles fundamental clauses of sentiment analysis, namely sentiment polarity distribution. This paper prospective a schema for online opinion representation. The inputs to the schema are product name, date and time and review of that product, where output contains the summary of the review in compact manner. The whole process includes summarization in three steps: (1) Product feature based, which is given by customer; (2) In each review, Identify expected features in each opinion sentence and (3) Finding out whether the feature/opinion is positive, negative or neutral and finally summary will be created [4]. This paper, propose a general method for finding opinion features from online reviews by exploiting the difference in opinion feature statistics across two compilation, an domain-specific corpus and one domain-independent corpus[5]. This product review paper discusses existing techniques and approaches for feature extraction in sentimental analysis and opinion mining [6]. This paper gives the feature based opinion mining and their efficiency in terms of precision, recall and accuracy. In this paper first extracts the feature, modifier and opinion.

## III. OPINION MINING AND SENTIMENT ANALYSIS

Sentiment analysis, also known as Opinion mining, is the study of sentiments that determines the judgement of people's opinions, sentiments, evaluations, and emotions in relation to entities such as products, services, organizations, events, topics and their different attributes. Generally opinion cannot structure a problem but it can subjective and in case opinion gathered from many people it should be summarized.

The approach of an opinion mining is described by [Jin.2006, Liu, 2010]. They put most impact on their work and finds the basic components of an opinion are:
Opinion holder: it is the person who gives a specific opinion on a particular object.
Object: it is called as the entity over which an opinion is expressed by user.
Opinion: it is a view, sentiment, or assessment of an object done by user.

Opinions are of two types described as: Regular and Comparative. Regular opinion is expressions on some target entities, which can be further classified into direct and indirect opinion. Comparative opinions are the Comparisons of more than one entity [7].

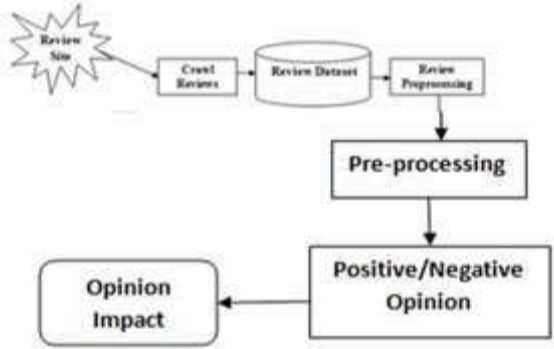

Fig. 1. Opinion Mining Work Flow

1. Direct opinions: This includes Text documents that give positive or negative opinion about the product directly. Example: "Battery backup of this mobile is too bad.
2. Comparison Opinions: Opinions found in text document and are meant to compare the object with some other objects. Opinion mining is the part of the web mining.

## IV. RESEARCH PATTERN AND METHODOLOGY

### A. Collection of Data

The appropriate amount of Data used in this paper is a arranged set of product reviews collected from amazon.com. From August to December 2018, in total, we collected over 500 sentiments of product reviews in which the products belong to 4 major categories: Mobiles, Computers, Flash drives and Electronics 3(a)). These online reviews were posted by over 3.2 millions of customers (reviewers) towards 10,001 products. Each review includes the following information: 1) reviewer_ID; 2) product model; 3) date and time of the review; 4) review text.

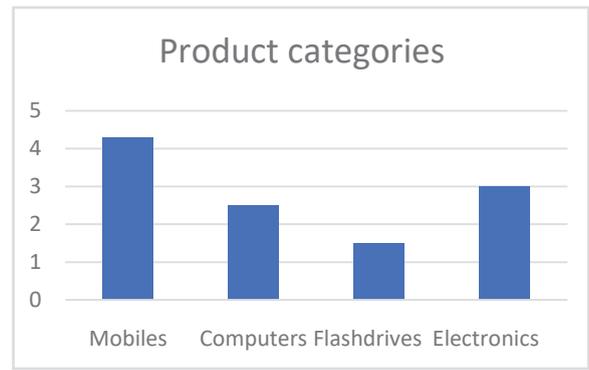

Fig. 2. Product Categories

### B. Sentiments Sentences Judgement and POS Tagging

This process is proposed by Pang and Lee [8] in which all objective content should be removed for analysis of sentiment. Instead of removing objective content, in our study, all subjective content was extracted for future analysis which consists of all sentiment sentences. A sentiment sentence is that which contains, at least, one positive or negative word. All of the sentences were first of all arranged into categorized English words. Every word of a sentence has its semantic role that defines how the word is used. The semantic roles are also called as the parts of speech. There are generally 8 parts of speech in English: the verb, the pronoun, the noun, some adverbs and prepositions, the interjection, and the conjunction. In natural language preference and suitability, part-of-speech (POS) taggers [9-10] have been generated to classify words based on their parts of speech. In sentiment classification, a POS tagger is most important because of the following two reasons: 1) Words like nouns and pronouns mostly do not contain any sentiment. So, it is able to filter out such words with the help of a POS tagger; 2) A POS tagger can also be useful in distinguish words that can be used in different parts of speech. For instance, as a verb, "improved" may conduct different amount of sentiment as being of an adjective. The POS tagger used for this survey is a max-entropy POS tagger developed for the Penn Treebank Project . This tagger is able to provide 46 different tags which indicate that it can identify more detailed semantic roles than only 8.

## V. CONCLUSION

Sentiment Analysis or opinion mining is a case study which analyses people's sentiments, attitudes, entropy or emotions towards certain entities. This paper tackles a fundamental problem of sentiment analysis, sentiment polarity categorization. The data for this research is collected of online product reviews from Amazon.com. A process known as sentiment polarity categorization and POS has been proposed along with detailed descriptions of each step. These steps consist of pre-processing, pre-filtering, biasing, data accuracy etc. features which require the knowledge of machine learning. A lot of work in opinion mining and sentiments of customer reviews has been conducted to mine opinions in form of document, sentence and feature level sentiment analysis. For future preferences, Opinion Mining can be carried out on set of discovered feature expressions extracted from reviews become a most interesting research area. There is more innovative and effective techniques have to be invented which should overcome the current challenges faced by Opinion Mining and Sentiment Analysis.